\documentclass[10pt,twocolumn,letterpaper]{article}

\usepackage{wacv}
\usepackage{times}
\usepackage{epsfig}
\usepackage{graphicx}
\usepackage{amsmath}
\usepackage{amssymb}
\usepackage{booktabs}
\usepackage{enumitem}
\usepackage{multirow}
\usepackage{subcaption}
\usepackage[breaklinks=true,letterpaper=true,colorlinks,bookmarks=false]{hyperref}
\usepackage{scrextend}

\DeclareGraphicsExtensions{.pdf,.jpg,.png,.jpeg}
\graphicspath{{images/}, {figs/}}

\newcommand{\secref}[1]{Section~\ref{sec:#1}}
\newcommand{\figref}[1]{Figure~\ref{fig:#1}}
\newcommand{\tabref}[1]{Table~\ref{tab:#1}}

\wacvfinalcopy

\def\wacvPaperID{493}

\ifwacvfinal\fi
\begin{document}

\title{FARSA: Fully Automated Roadway Safety Assessment}

\author{
  Weilian Song$^1$\hspace{1cm}Scott Workman$^1$\hspace{1cm}Armin Hadzic$^1$\hspace{1cm}\hspace{1cm}Xu Zhang$^{2,3}$\\
  Eric Green$^{2,3}$\hspace{1cm}Mei Chen$^{2,3}$\hspace{1cm}Reginald Souleyrette$^{2,3}$\hspace{1cm}Nathan Jacobs$^1$\\
  \\[.2cm]
  \begin{minipage}{\linewidth}
    \centering
    $^1$Department of Computer Science, University of Kentucky\\
    $^2$Department of Civil Engineering, University of Kentucky\\
    $^3$Kentucky Transportation Center, University of Kentucky\\
  \end{minipage}
}

\maketitle
\ifwacvfinal\thispagestyle{empty}\fi

\begin{abstract}

  This paper addresses the task of road safety assessment. An 
  emerging approach for conducting such assessments in the United
  States is through the US Road Assessment Program (usRAP), which rates
  roads from highest risk (1 star) to lowest (5 stars).
  Obtaining these ratings requires manual, fine-grained labeling of
  roadway features in street-level panoramas, a slow and costly
  process. We propose to automate this process using a deep
  convolutional neural network that directly estimates the star rating
  from a street-level panorama, requiring milliseconds per image at
  test time. Our network also estimates many other road-level
  attributes, including curvature, roadside hazards, and the type of
  median. To support this, we incorporate task-specific attention
  layers so the network can focus on the panorama regions that are
  most useful for a particular task.  We evaluated our approach on a
  large dataset of real-world images from two US states. We found that
  incorporating additional tasks, and using a semi-supervised training
  approach, significantly reduced overfitting problems, allowed us to
  optimize more layers of the network, and resulted in higher
  accuracy.

\end{abstract}

\section{Introduction}

At over 35,000 fatalities annually, highway crashes are one of the
primary causes of accidental deaths in the US~\cite{xu2016deaths}.
Data-driven approaches to highway safety have been widely used to
target high-risk road segments and intersections through various
programs leading to a reduction in the number of fatalities
observed~\cite{wu2012evaluation}. Unfortunately, this reduction has plateaued in
recent years.  The prospect of automated vehicles will likely have a
dramatic effect on further reducing fatalities---perhaps as much as
90\%~\cite{gibson2017analysis}. However, until this technology is
mature, policies are in place, and the public adopts machine-driven
vehicles, we must rely on other methods to improve safety. In the
USA and other developed countries, high crash-rate locations have been
predominantly identified and addressed.  State and local highway
authorities are now turning increasingly to systemic analysis to
further drive down the occurrence of highway crashes, injuries, and
fatalities.  Systemic analysis requires data to identify and assess
roadway features and conditions known to increase crash risk. 

We address the task of automatically assessing the safety of roadways
for drivers. Such ratings can be used by
highway authorities to decide, using cost/benefit analysis, where to
invest in infrastructure improvements. Automating this typically
manual task will enable much more rapid assessment of safety problems
of large regions, entire urban areas, counties, and states. In the
end, this can save lives and reduce injuries.

\begin{figure}
  \centering
  \includegraphics[height=.37\linewidth]{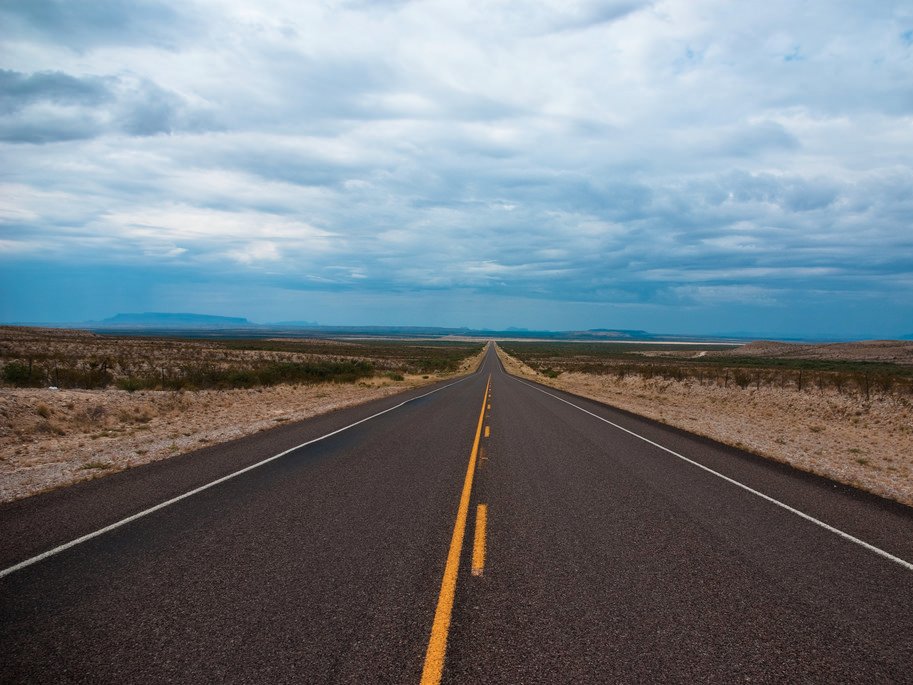}
  \includegraphics[height=.37\linewidth]{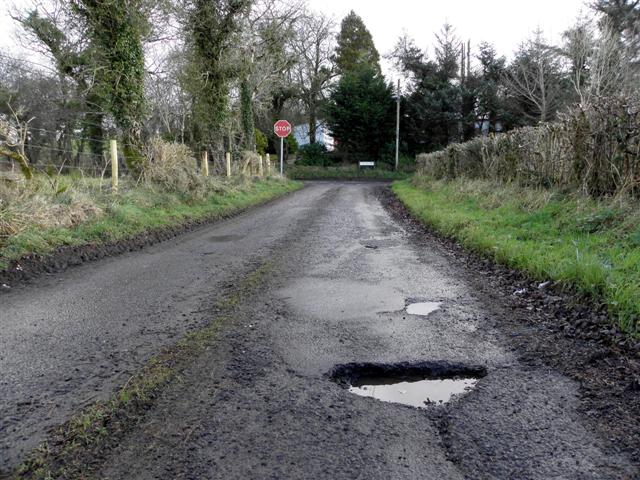}
  \includegraphics[height=.37\linewidth]{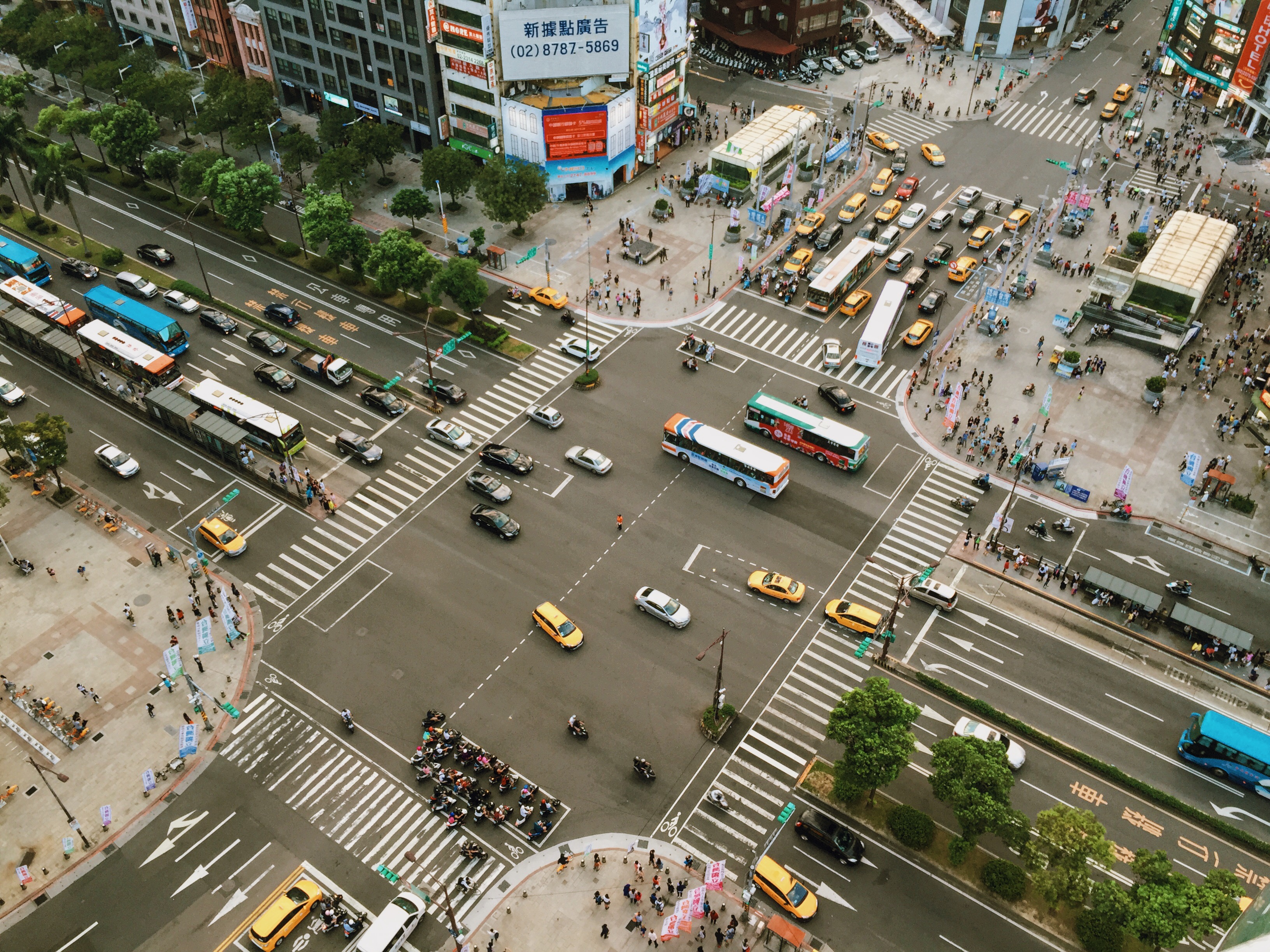}
  \includegraphics[height=.37\linewidth]{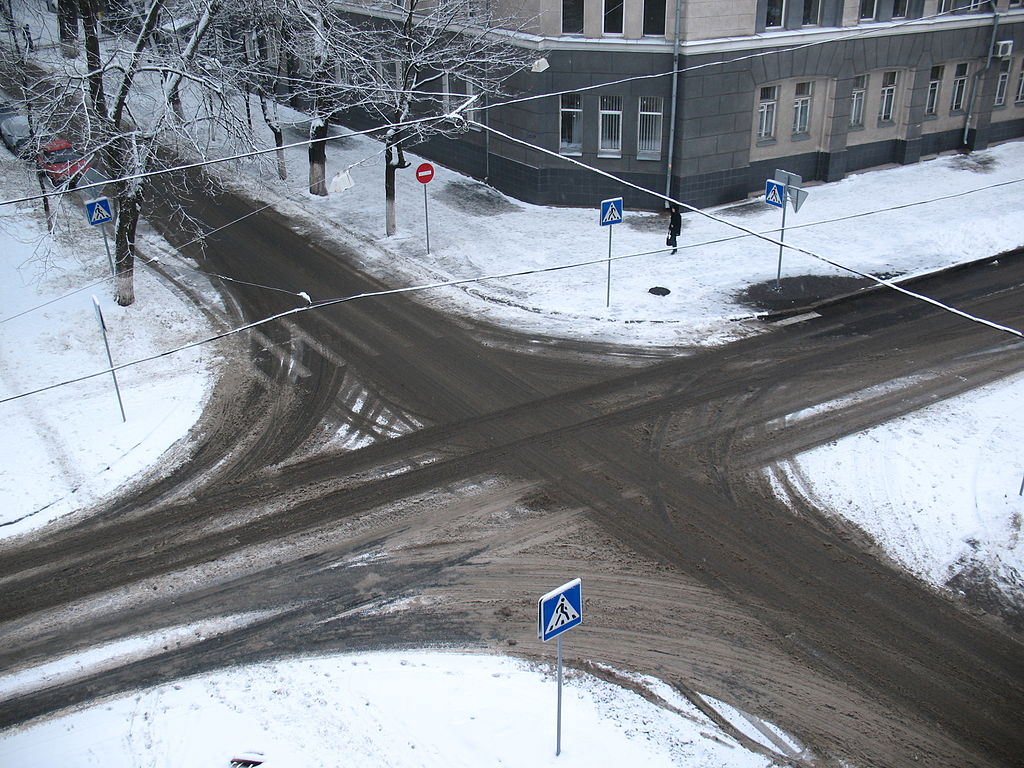}
  \caption{Some roads are safer than others. Our work explores
  automatic methods for quantifying road safety.}
  \label{fig:cartoon}
\end{figure}

An emerging technique for systemically addressing the highway crash
problem is represented by the protocols of the United States Road
Assessment Program (usRAP). In this process, a trained coder
annotates, at regular intervals, various features of the roadway such
as roadway width, shoulder characteristics and roadside hazards, and
the presence of road protection devices such as guardrail. These
annotations are based on either direct observation or from imagery
captured in the field and can be used to augment any existing highway
inventory of these features. These data are then used to rate the
roadway, and in turn, these ratings may be classified into a 5-tier
star-rating scale, from highest risk (1 star) to lowest (5 stars). This
manual process is laborious and time consuming, and sometimes cost
prohibitive. Moreover, the speed and accuracy of the rating process
can and does vary across coders and over time.

\begin{figure}

  \centering

  \includegraphics[width=.98\linewidth]{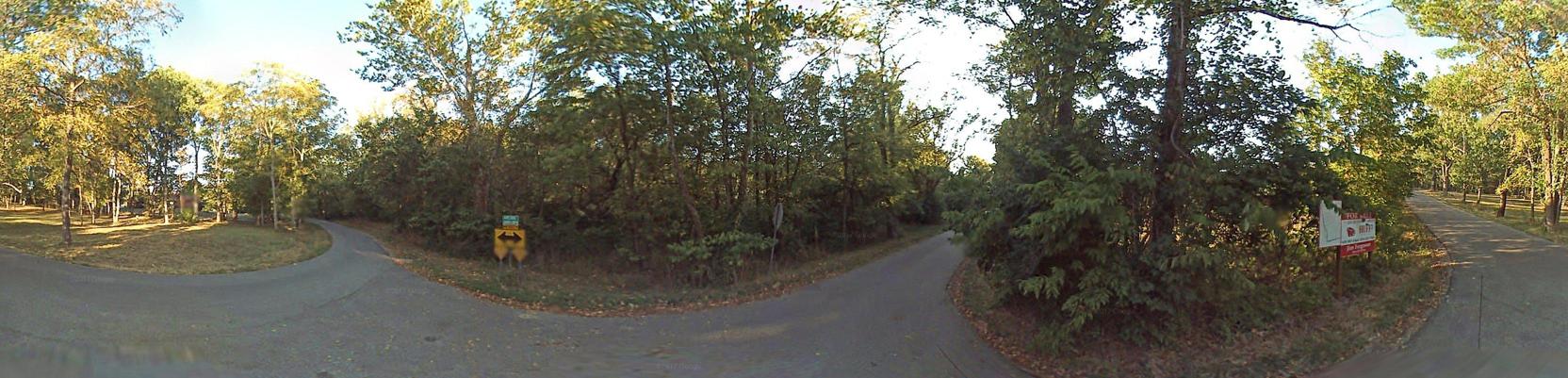}
  \includegraphics[width=.98\linewidth]{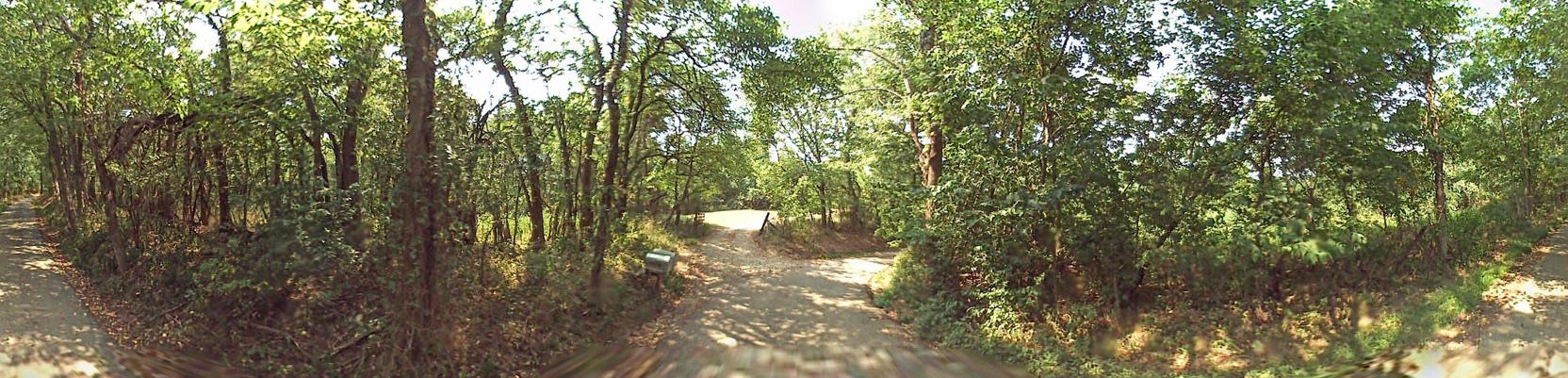}
  \includegraphics[width=.98\linewidth]{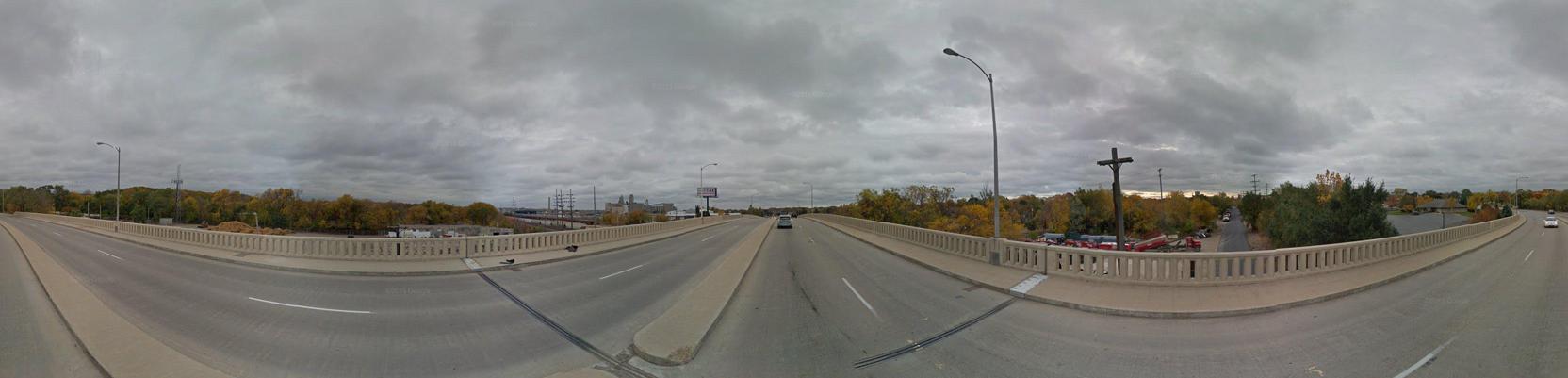}
  \includegraphics[width=.98\linewidth]{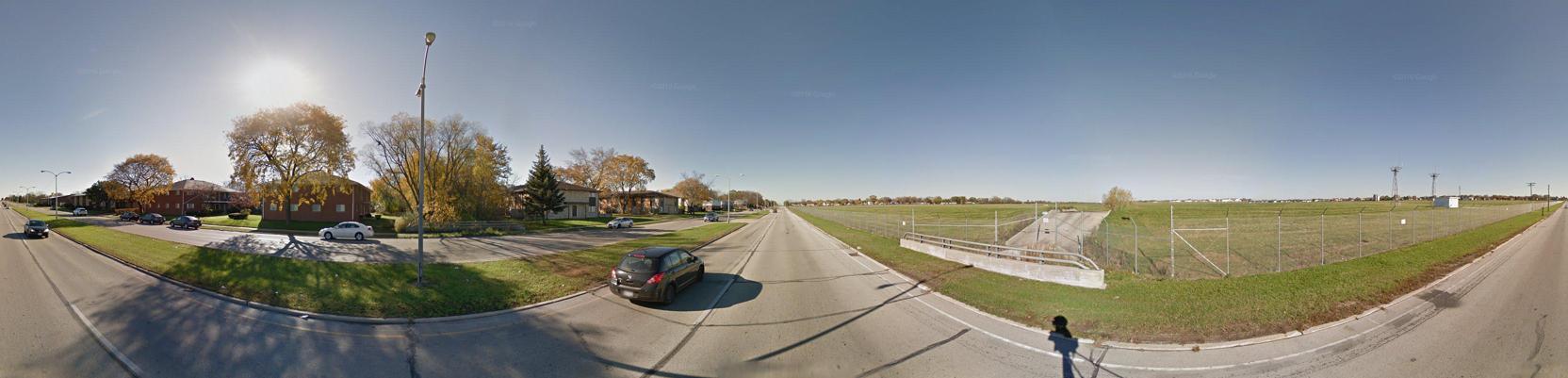}

  \caption{Example panoramas that our system uses to estimate the
  safety star rating. The top two are rated 1 star, the most
dangerous.  The bottom two are rated 5 star, the safest. }

  \label{fig:examples}

\end{figure}

To automate this manual process, we propose a deep convolutional
neural network (CNN) architecture that directly estimates the star
rating from a ground-level panorama.  See \figref{examples} for
examples of input panoramas.  Note the lack of a physical median,
paved shoulder, and sidewalks in the less safe roads. The key features
of our approach are:
\begin{itemize}

    \item {\em multi-task learning}: we find that augmenting the
      network to support estimating lower-level roadway features
      improves performance on the main task of estimating the star
      rating; 
    
    \item {\em task-dependent spatial attention layer}: this allows
      the network to focus on particular regions of the panorama that
      are most useful for a particular task; and

    \item {\em unsupervised learning}: we add a loss that
      encourages the star rating distribution to be similar for nearby
      panoramas, greatly expanding the number of panoramas seen by the
      network, without requiring any manual annotation.

\end{itemize}
We evaluate this approach on a large dataset of real-world imagery and
find that it outperforms several natural baselines. Lastly, we present
an ablation study that demonstrates the benefits of the various
components of our approach.

\section{Related Work}

Our work builds upon work in several areas: general purpose scene
understanding, automatic methods for understanding urban areas, and
current practice in the assessment of roadway features.

\paragraph{Scene Classification and Image Segmentation}

Over the past ten years, the state of the art in scene classification
has dramatically improved.  Today, most, if not all, of the best
performing methods are instances of deep convolutional neural networks
(CNNs).  For many tasks, these methods can estimate the probability
distribution across hundreds of classes in milliseconds at human-level
accuracy~\cite{he2015delving}.  A notable example is the {\em Places}
CNN~\cite{zhou2014learning}, developed by Zhou et al., which adapts a
network that was created for image
classification~\cite{krizhevsky2012imagenet}.  This network, or
similar networks, has been adapted for a variety of tasks, including:
horizon line estimation~\cite{zhai2016context}, focal length
estimation~\cite{workman2015deepfocal},
geolocalization~\cite{lin2015learning}, and a variety of
geo-informative attributes~\cite{lee2015predicting}.

This ability to adapt methods developed for scene classification and
image segmentation to other tasks was one of the main motivations for
our work.  However, we found that na\"ively applying these techniques
to the task of panorama understanding did not work well. The main
problem is that these methods normally use lower resolution
imagery which means they cannot identify small features of the roadways
that have a significant impact on the assessment of safety.  We
propose a CNN architecture that overcomes these problems by
incorporating a spatial attention mechanism to support the extraction
of small image features.
 
In some ways, what we propose is similar to the task of
semantic segmentation~\cite{cordts2016cityscapes}, which focuses on
estimating properties of individual pixels. The current best methods
are all CNN-based and require a densely annotated training
set. Constructing such datasets is a labor intensive process.
Fortunately, unlike semantic segmentation, we are estimating a
high-level attribute of the entire scene so the effort required to
construct an annotated dataset is lower. It also means that our CNNs
can have a different structure; we extract features at a coarser
resolution and have many fewer loss computations. This means faster
training and inference. 

\paragraph{Urban Perception and High-Definition Mapping}

Recently there has been a surge in interest for applying techniques
for scene
classification~\cite{arietta2014city,dubey2016deep,naik2014streetscore,ordonez2014learning,salesses2013collaborative,workman2017natural,workman2017unified}
and image segmentation to understanding urban areas and transportation
infrastructure~\cite{geiger2012we,mattyus2016hd,wang2017torontocity}.
The former focuses on higher-level labels, such as perceived safety,
population density, or beauty, while the later focuses on
finer-grained labels, such as the location on line markings or the
presence of sidewalks.  Our work, to some extent, combines both of
these ideas.  However, we focus on estimating the higher-level labels
and use finer-grained labels as a form of multi-task learning. Our 
evaluation demonstrates that by combining these in a single
network enables better results.
    
\paragraph{Current Practice in Roadway Assessment}

The Highway Safety Manual (HSM)~\cite{national2010highway} outlines
strategies for conducting quantitative safety analysis, including a
predictive method for estimating crash frequency and severity from
traffic volume and roadway characteristics. The usRAP Star Rating
Protocol is an internationally established methodology for assessing road safety, and
is used to assign road protection scores resulting in five-star ratings of road segments. A star
rating is partly determined based on the presence of approximately 60
road safety features~\cite{harwood2010validation}.  More
implementation of road safety features entails a higher safety rating,
and vice versa. There are separate ratings for vehicle occupants,
cyclists, and pedestrians, but we focus on vehicle occupant star
ratings in this paper.

\section{Approach}

We propose a CNN architecture for automatic road safety assessment.
We optimize the parameters of this model by minimizing a loss function
that combines supervised, multi-task, and unsupervised component
losses. We begin by outlining our base architecture, which we use in
computing all component loss functions.

\subsection{Convolutional Neural Network Architecture}
\label{sec:base}

Our base CNN architecture takes as input a street-level,
equirectangular panorama (\eg, from Google Street View) and outputs a
categorical distribution over a discrete label space.  Our focus is on
the roadway safety label space, which is defined by usRAP to have five
tiers. Other label spaces will be defined in the following section.
In all experiments, panoramas are cropped vertically, to reduce the
number of distorted sky and ground pixels, and then resized to be $224
\times 960$. 

\begin{figure*}
	\centering
    \includegraphics[width=\textwidth, height=2in]{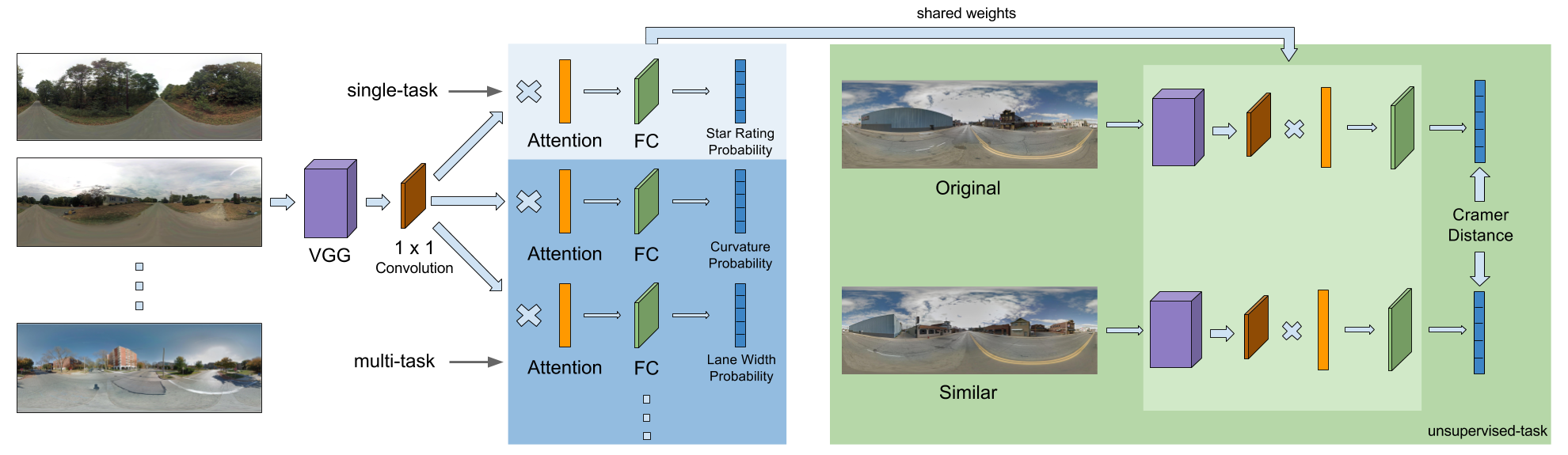}
    \caption{Overview of our network architecture.}
    \label{fig:architecture}
\end{figure*}

The CNN consists of a portion of the VGG
architecture~\cite{simonyan2014vgg}, followed by a $1\times1$
convolution, a spatial attention layer, and a final fully connected
layer. See \figref{architecture} for a visual overview of the full
architecture and the various ways we use it to train our model.  
We use the VGG-16 architecture~\cite{simonyan2014vgg} for low-level
feature extraction. It consists of 5 convolutional blocks followed by
3 fully connected layers, totaling 16 individual layers. We remove the
fully connected layers and use the output of the last convolutional
block, after spatial pooling. We denote this output feature map as
$S_1$, which is a $7 \times 30$ tensor with 512 channels.  $S_1$ is
then passed to a convolutional layer (ReLU activation function), with
kernels of size $1\times1$, a stride of 1, and 512 output channels.
The resulting tensor, $S_2$, has the following shape: $7 \times 30
\times 512$. This gives us a set of mid-level features that we use to
predict the safety of a given panorama.

As the target safety rating may depend on where in the image a
particular roadway feature is present, we introduce an attention
layer to fuse the mid-level features.  Specifically, we use a
learnable vector $\boldsymbol{a}$ that takes the
weighted average of $S_2$ across the first two dimensions. Our process
is as follows: we flatten the spatial dimensions of $S_2$ ($210 \times
512$) and multiply it by $\boldsymbol{a}$ ($1 \times 210$). This
process is akin to global average pooling~\cite{lin2014network}, but
with location-specific weights. The output ($1 \times 512$) is then
passed to a task-dependent fully connected layer with $K$
outputs.

\subsection{Loss Function}
\label{sec:loss}

A key challenge in training large neural network architectures,
especially with small datasets, is avoiding overfitting. For this
work, we propose to use a combination of supervised, multi-task, and
unsupervised learning to train our model. The end result is a total
loss function:
\begin{equation} \nonumber
L = \lambda_s L_s + \lambda_m L_m + \lambda_u L_u.
\end{equation}
Each component loss function processes panoramas using the base architecture
defined in the previous section. In all cases, the parameters of the VGG
sub-network and the subsequent $1\times1$ convolution are tied.  The attention
and final fully connected layers are independent and, therefore, task specific. The
remainder of this section describes the various components of this loss in
detail.

\subsubsection{Supervised Loss}
\label{sec:supervised}

The first component, $L_s$, of our total loss, $L$, corresponds to the
main goal of our project: estimating the safety of a given roadway from an input
panorama. Each panorama is associated with a star rating label, $l \in
\{1,\ldots,5\}$. We apply the network defined in \secref{base} with
$K=5$ outputs, representing a categorical distribution over star
ratings. Our first loss, $L_s$, incorporates both a classification and
regression component. For classification, we use the standard cross-entropy loss
between the predicted distribution, $\hat{y}$, and target distribution, $y$:
\begin{equation} \label{eq:1}
  L_{s^1} = - \frac{1}{N} \sum^{N}_{i=1} y_{i}(l_i) \log \hat{y}_{i}(l_i)
\end{equation}
where $N$ is the number of training examples. For regression, we use the
Cramer distance between $\hat{y}$ and $y$:
\begin{equation} \label{eq:2}
  L_{s^2} = \frac{1}{N} \sum^{N}_{i=1} \| F(\hat{y}_{i}) - F(y_{i}) \|_2^2
\end{equation}
where $F(x)$ is the cumulative distribution function of $x$. Each component is
weighted by $\lambda_{s^1}$ and $\lambda_{s^2}$, respectively:
\begin{equation}
  L_s = \lambda_{s^1} L_{s^1} + \lambda_{s^2} L_{s^2}.
\end{equation}

\subsubsection{Multi-task Loss}
\label{sec:multitask}

The second component, $L_m$, of our total loss, $L$, represents a set of
auxiliary tasks.  We selected M auxiliary tasks with discrete label space for
learning, specifically: area type, intersection channelization, curvature,
upgrade cost, land use (driver/passenger-side), median type, roadside severity
(driver/passenger-side distance/object), paved shoulder
(driver/passenger-side), intersecting road volume, intersection quality,
number of lanes, lane width, quality of curve, road condition, vehicle
parking, sidewalk (passenger/driver-side), and facilities for bicycles.
All images were annotated by a trained coder as part of the usRAP
protocol. 

For each new task, the prediction process is very similar to the
safety rating task, with its own attention mechanism and final
prediction layer.  The only difference is the output size of the
prediction layer which varies to match the label space of the specific
task. To compute $L_m$, we sum the cross-entropy loss across all
tasks: $L_m = \sum^{M}_{t=1} L_t$ where $L_t$ is the loss for task $t$.

\subsubsection{Unsupervised Loss} 
\label{sec:unsupervised}

The third component, $L_u$, of our total loss, $L$, represents Tobler's First
Law of Geography: ``Everything is related to everything else, but near things
are more related than distant things.'' Specifically, we assume that
geographically close street-level panoramas should have similar star ratings,
and we encourage the network to produce identical output distributions for adjacent
panoramas. While this assumption is not always true, we find that it
improves the accuracy of our final network.

The key feature of this loss is that it does not require the
panoramas to be manually annotated. Therefore, we can greatly expand our
training set size by including unsupervised examples.  This is important,
because due to the small size of the safety rating dataset and a large number
of parameters in the network, we found it impossible to update VGG layer
weights without overfitting when using only supervised losses.

To define this loss, we first build a set, $U = \{(a_i,b_i)\}$ of panorama
pairs. The pairs are selected so that they are spatially close (within 50
feet) and along the same road. For each pair, we compute the Cramer distance
between their rating predictions, $(\hat{y}_{a}, \hat{y}_{b})$, as in:
\begin{equation}
  L_{u} = \frac{1}{|U|} \sum_{(a,b) \in U} \| F(\hat{y}_a) - F(\hat{y}_b) \|_2^2.
\end{equation}

\subsection{Implementation Details}

Our network is optimized using ADAM~\cite{kingma2014adam}. We
initialize VGG using weights pre-trained for object
classification~\cite{russakovsky2015imagenet}. We experimented with
weights for scene categorization~\cite{zhou2017places} and found that
object classification was superior, and both were significantly better
than random initialization. We allow the final convolutional block of VGG to optimize
with learning rate 0.0001, while all task-specific layers use a
learning rate of 0.001.  We decay the learning rates three times during
training, by a factor of 10 each time.

Through experimentation, we find that optimizing the total loss with
$\lambda_{s^1}=1$, $\lambda_{s^2}=100$, $\lambda_s=0.1$, $\lambda_m=1$, and $\lambda_u=0.001$ offers the best
results. We use ReLU activations throughout, except for attention mechanisms
and final output layers, which apply a softmax function.  For every trainable
layer other than the attention mechanisms, we apply $L_2$ regularization to
the weights with a scale of 0.0005. We train with a mini-batch size of 16,
with each batch consisting of 16 supervised panoramas and labels along with 16
pairs of unsupervised panoramas.

\section{Evaluation}

We evaluate our methods both quantitatively and qualitatively. Below
we describe the datasets used for these experiments, explore the
performance of our network through an ablation study, and present an
application to safety-aware vehicle routing.

\subsection{Datasets}

For supervised and multi-task training, we utilize a dataset annotated through
the U.S. Road Assessment Program (usRAP), which contains star ratings and
auxiliary task labels for 100-meter segments of roads in both {\em Urban} and
{\em Rural} regions. To obtain the labels for each location, a trained coder
visually inspects the imagery for a road segment and assigns multi-task labels
using a custom user interface. During this process the coder is free to adjust
the viewing direction. The star rating for each location is then calculated
from the auxiliary labels using the Star Rating Score equations. For more
information on dataset annotation methods, please refer to \cite{usrap}
\cite{harwood2010validation} \cite{nye2017usrap}.

\begin{figure}
    \centering
    \includegraphics[width=.48\linewidth,height=2.7in]{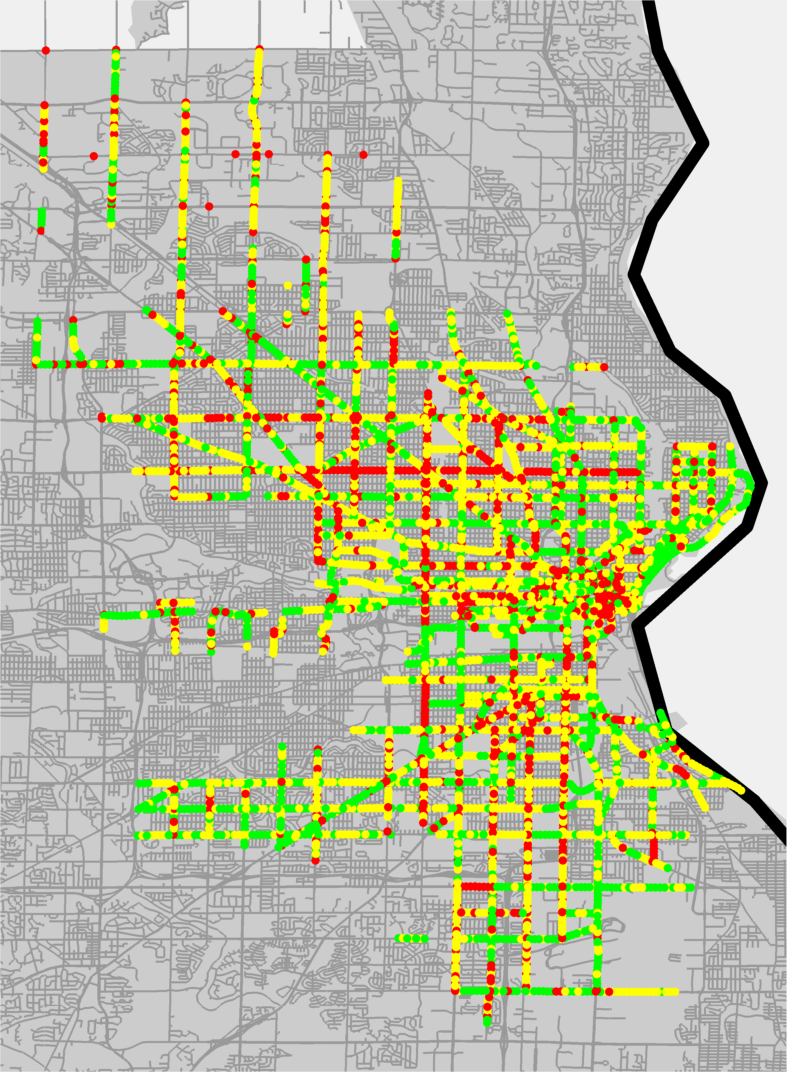}
    \includegraphics[width=.48\linewidth,height=2.7in]{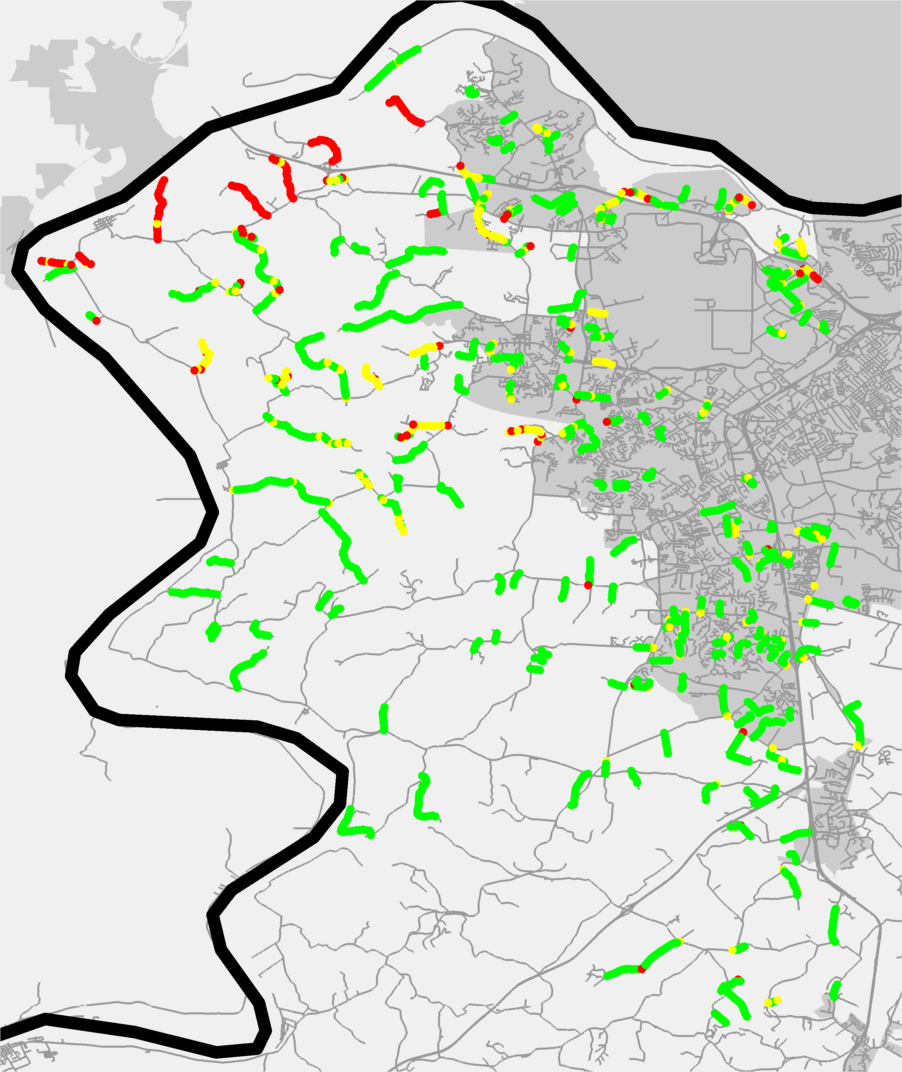}
    \caption{Maps showing the locations and star ratings (red is
      highest risk, yellow is average risk, and green is least risk) of the
      panoramas in our {\em Urban} (left) and {\em Rural} (right)
    study regions.}
    \label{fig:scatter}
\end{figure}

The {\em Rural} area has 1,829 panoramas and the {\em Urban} area has 5,459
panoramas, for a total of 7,288 samples. \figref{scatter} shows scatter plots
of panorama locations for each region, color coded by the safety score manually
estimated from the imagery.
For unsupervised training data, we uniformly sample road segments from
the state surrounding the {\em Rural} region and then query for
panoramas that are less than 50-feet apart along the segment. The
result is a dataset of approximately 36,000 pairs of panoramas.

\subsection{Preprocessing}
\paragraph{Panorama Processing} \label{preprocessing}
For each sample, we download an orbital panorama through Google Street View.
We orient panoramas to the heading used during manual coding.  This is important because safety
ratings are sensitive to the direction of travel. During training only, random
augmentation is performed by randomly jittering the direction of travel
uniformly between -5 and 5 degrees. Finally, each image is cropped vertically
and reshaped to $224\times960$. The cropping operation removes the unneeded sky and
ground portion of the panorama, and we preserve the aspect ratio of the
cropped image when reshaping.

\paragraph{Train/Test Split}
To create train/test splits for network training and evaluation, we utilize
a stratified sampling method for each region's data, 90\% for train and 10\%
for test. A validation set (2\% of train) is used for model selection.
Corresponding splits from the two regions are combined to form the
final sets. We have ensured that locations in the test set are at
least 300 meters away from locations in the train set.

\paragraph{Class Weighting}
The distribution of labels in our training split is very unbalanced, which led
to poor fitting to panoramas with labels in the minority groups (1 star roads
specifically). To alleviate this issue, we deploy a class weighting strategy on
the star rating loss function to proportionally adjust the loss of each sample
based on the frequency of its ground truth label.

We first find the weight vector of each star rating class $w$ through the
equation below:
\begin{equation}
  w_l = \frac{N}{K * count(l)}
\end{equation}
where $w_l$ is the weight for class $l$. With the weight vector $w$, we modify
equation \ref{eq:1} and \ref{eq:2} as follows,
\begin{equation}
  L_{s^1} = - \frac{1}{N} \sum^{N}_{i=1} w_{l_i} y_{i}(l_i) \log \hat{y}_{i}(l_i)
\end{equation}
\begin{equation}
  L_{s^2} = \frac{1}{N} \sum^{N}_{i=1} w_{l_i} \| F(\hat{y}_{i}) - F(y_{i}) \|_2^2
\end{equation}
where $w_{l_i}$ is the weight for class $l_i$.
\begin{table}[]
  \centering
  \caption{Parameter settings for each method.}
  \begin{tabular}{lcclcl}
    \toprule
    Method   & $\lambda_{s^1}$ & $\lambda_{s^2}$ & $\lambda_s$ & $\lambda_m$ & $\lambda_u$ \\ \hline
    {\em Baseline} &       1       &     $10^2$      & 1          & 0           & 0           \\
    {\em M1}       &       1       &     $10^2$      & 1          & 0           & 0           \\
    {\em M2}       &       1       &     $10^2$      & 0.1        & 1           & 0           \\
    {\em M3}       &       1       &     $10^2$      & 1          & 0           & 0.001      \\
    {\em M4}       &       1       &     $10^2$      & 0.1        & 1           & 0           \\
    {\em Ours}     &       1       &     $10^2$      & 0.1        & 1           & 0.001      \\
    \bottomrule
  \end{tabular}
  \label{tab:methods_table}
\end{table}
\subsection{Ablation Study}

We compare our complete architecture with five variants. The simplest
method, {\em Baseline}, omits the adaptive attention layer and instead
equally weights all parts of the input feature map (i.e., global
average pooling).  The remaining variants include different
combinations of adaptive attention, multi-task loss, and unsupervised
loss.  For each method, we performed a greedy hyperparameters search
using our validation set.  We initially selected the optimal weighting
of $\lambda_{s^2}$ relative to $\lambda_{s^1}$ by selecting
$\lambda_{s^2}$ from $(.01,.1,1,10,100)$. We used the same strategy
when adding a new feature, while keeping the existing hyperparameters
fixed.  We train all networks for 10,000 iterations.

\begin{table}[]
  \centering
  \caption{Top-1 accuracy for each method.}
  \begin{tabular}{lcccc}
    \toprule
    Method   & Attn.     & Multi.      & Unsuper. & Acc.  \\ \hline
    {\em Baseline} &           &           &          & 43.06 \\
    {\em M1}       &     X     &           &          & 43.28 \\
    {\em M2}       &           &     X     &          & 43.56 \\
    {\em M3}       &     X     &           &     X    & 43.63 \\
    {\em M4}       &     X     &     X     &          & 45.68 \\
    \hline
    {\em Ours}     &     X     &     X     &     X    & \bf{46.91} \\
    \bottomrule
  \end{tabular}
  \label{tab:eval}
\end{table}

\tabref{eval} displays the macro-averaged accuracy for all methods on
the test set, where for each method we compute the per-class accuracy
and average the class accuracies. Our method outperforms all other
methods, each of which includes a subset of the complete approach.  We
observe that the {\em Baseline}, {\em M1}, {\em M2}, and {\em M3}
methods all achieve similar accuracy. The next method, {\em M4}, which
combines the per-task attention and multi-task learning, performs
significantly better. It seems that the multi-task loss and attention
in isolation are not particularly helpful, but when they are combined
they lead to improved performance.  We also observe that the
unsupervised loss is only significantly helpful when combined with
per-task attention and the multi-task loss. 

\begin{figure}
  \centering
    \includegraphics[width=\linewidth]{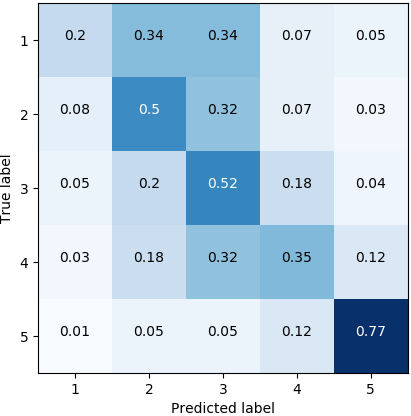}
    \caption{The test-set confusion matrix (row normalized) for our method.}
    \label{fig:confusion}
\end{figure}

Figure \ref{fig:confusion} shows the test-set confusion matrix for our best method. The
results are quite satisfactory except for the 1-star rating category, but that
is as expected due to the imbalanced nature of our training dataset, with only
5.6\% of all samples being 1-star roads. 

\subsection{Visualizing Attention}

\begin{figure}

  \centering
  
  \begin{subfigure}{1\linewidth}
    \includegraphics[width=1\linewidth]{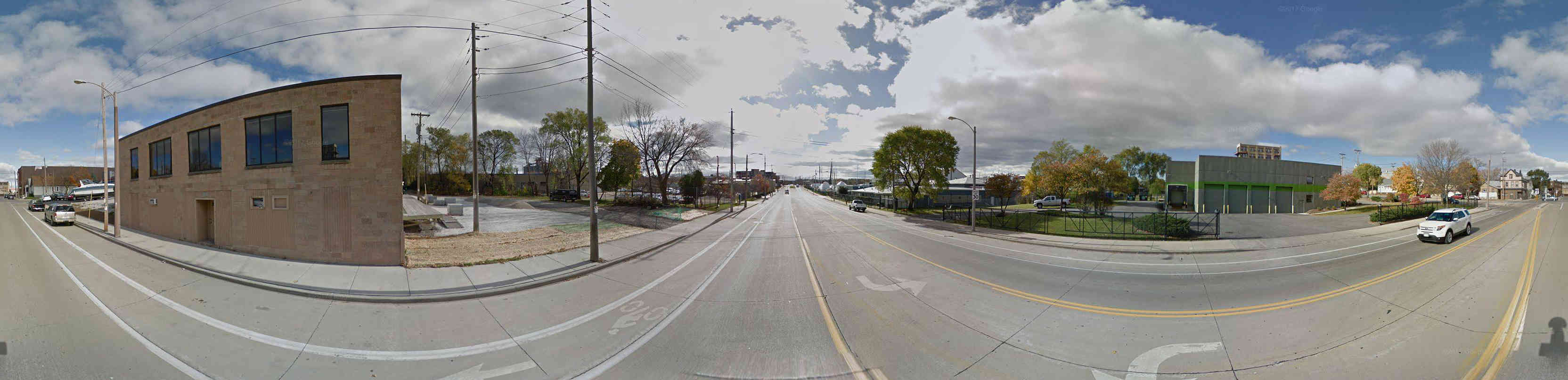}
    \caption*{Example Panorama}
  \end{subfigure}
  \begin{subfigure}{1\linewidth}
    \includegraphics[width=1\linewidth]{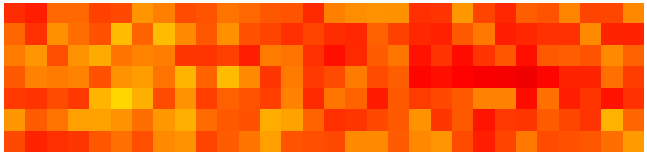}
    \caption{Star Rating}
  \end{subfigure}
  \begin{subfigure}{1\linewidth}
    \includegraphics[width=1\linewidth]{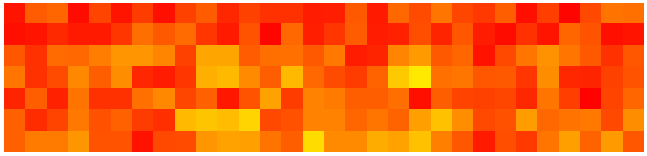}
    \caption{Median Type}
  \end{subfigure}
  \begin{subfigure}{1\linewidth}
    \includegraphics[width=1\linewidth]{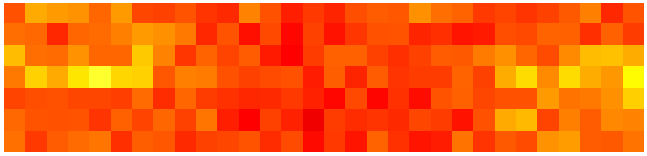}
    \caption{Land Use (driver-side)}
  \end{subfigure}
  \begin{subfigure}{1\linewidth}
    \includegraphics[width=1\linewidth]{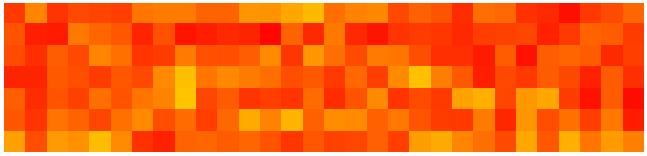}
    \caption{Number of Lanes}
  \end{subfigure}
  \begin{subfigure}{1\linewidth}
    \includegraphics[width=1\linewidth]{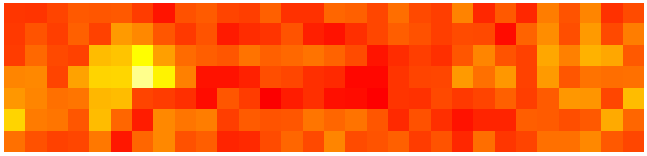}
    \caption{Roadside Severity (driver-side object)}
  \end{subfigure}

  \caption{Heatmaps that visualize the weights of the task-specific
  attention layers. White (red) corresponds to higher (lower)
  weight.}

  \label{fig:attention}

\end{figure}

As described in \secref{multitask}, each task has an independent
attention mask, $\boldsymbol{a}$, whose shape ($1 \times 210$)
corresponds to the flattened spatial dimensions of the feature map,
$S_2$, output by the previous convolutional layer.  Therefore, a
reshaped version of $\boldsymbol{a}$ corresponds to the regions in
$S_2$ that are important when making predictions for a given task.
\figref{attention} visualizes the attention mechanism for our main
task and several of our auxiliary tasks, where lighter (yellow) regions
have higher weight than darker (red) regions. For example, in
\figref{attention} (e), attention is focused on the left side of the
panorama, which makes sense given the task is related to identifying
dangerous driver-side objects, such as a drainage ditch or safety
barrier.

\subsection{Multi-task Evaluation}

We evaluate the multi-task prediction performance of our architecture.
\tabref{multi} shows the Top-1 macro-averaged accuracy for each of our
auxiliary tasks, along with random performance if we sample predictions based
on the prior distribution of labels for that task. The rightmost column shows
the relative increase in accuracy for each task over random
performance, in some cases with a performance gain of almost 50\%.
\begin{table}[]
  \centering
  \caption{Multi-task evaluation for our architecture.  
  R. = roadside, P. = passenger, and D. = driver.}
  \label{tab:multi}
  \begin{tabular}{llll}
    \toprule
    Label type                & Top-1 & Random & \% inc. \\
    \hline
    Area type                 & 97.40 & 50.39  & 47.01   \\
    Lane width                & 77.22 & 33.25  & 43.97   \\
    Curve quality             & 63.57 & 32.78  & 30.79   \\
    P. side land use          & 54.88 & 28.38  & 26.50   \\
    D. side land use          & 52.72 & 28.32  & 24.40   \\
    D. side sidewalk          & 77.17 & 57.00  & 20.17   \\
    Vehicle parking           & 52.39 & 33.06  & 19.33   \\
    Road condition            & 51.38 & 33.35  & 18.03   \\
    P. side sidewalk          & 60.34 & 43.54  & 16.80   \\
    Intersection quality      & 83.06 & 66.69  & 16.37   \\
    Intersection road volume  & 28.21 & 14.18  & 14.03   \\
    D. side paved shoulder    & 38.61 & 24.71  & 13.90   \\
    P. side paved shoulder    & 37.56 & 24.42  & 13.14   \\
    Number of lanes           & 45.78 & 33.13  & 12.65   \\
    R. D. side distance       & 37.51 & 25.47  & 12.04   \\
    R. P. side distance       & 36.45 & 24.98  & 11.47   \\
    Median type               & 57.30 & 46.86  & 10.44   \\
    R. P. side object         & 38.89 & 29.59  &  9.30   \\
    Upgrade cost              & 42.82 & 33.93  &  8.89   \\
    R. D. side object         & 49.24 & 41.15  &  8.09   \\
    Intersect channel         & 56.25 & 49.88  &  6.37   \\
    Bicycle facilities        & 75.19 & 71.51  &  3.68   \\
    Curvature                 & 25.83 & 25.52  &  0.31   \\
    \bottomrule
  \end{tabular}

\end{table}
\subsection {Safety-Aware Vehicle Routing} 
While our focus is on road safety assessment for infrastructure
planning, our network could be used for other tasks. Here we show how
it could be used to suggest less-risky driving routes.
We use a GPS routing engine that identifies the optimal path, usually
the shortest and simplest, a vehicle should traverse to
reach a target destination.  Some work has been done to explore semantic
routing with respect to scenery~\cite{DBLP:journals/corr/QuerciaSA14}, carpooling~\cite{He2014IntelligentCR}, and personalization~\cite{Sha:2013:SVN:2444776.2444798}. We propose routing that employs road safety scores for navigating to a
destination, in order to balance speed and safety. 

We selected a subset of the panorama
test split and used it to influence the Mapzen Valhalla routing engine’s edge
cost calculation. From the subset, each safety score was used as an edge
factor corresponding to the safety score’s GPS coordinate. When the routing
engine searched the GPS graph for a route’s traversal cost, it would identify
if a cost factor corresponding to a specific edge existed in the subset. Should a cost factor be present, the edge cost of the traversal would be $c_{edge}=c_{o}*factor$. The routing engine utilizes the augmented edge costs to determine the optimal route, namely, the lowest cumulative cost route. 

\begin{figure}
  \centering

  \includegraphics[width=.49\linewidth]{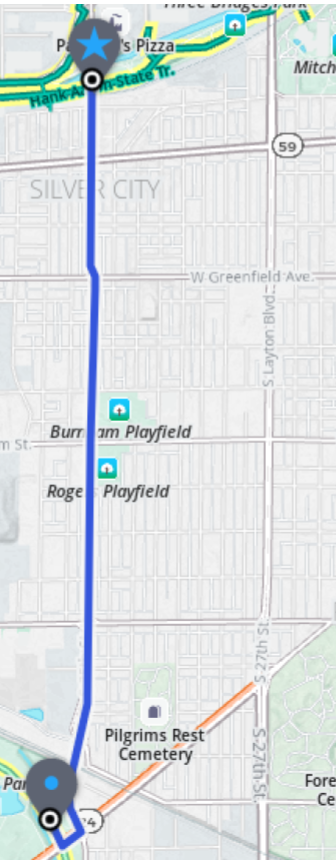}
  \hfill
  \includegraphics[width=.49\linewidth]{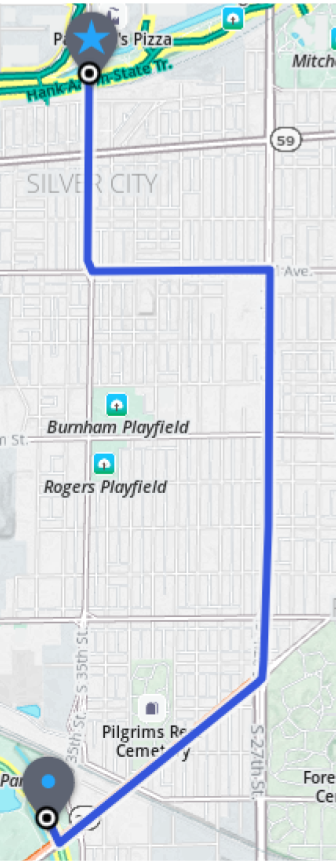}
  
  \caption{Unaltered routing (left) and predicted safety score route (right).}

  \label{routes}

\end{figure}

\begin{figure}
  \centering

  \begin{subfigure}{\linewidth}
    \centering
    \includegraphics[width=\linewidth]{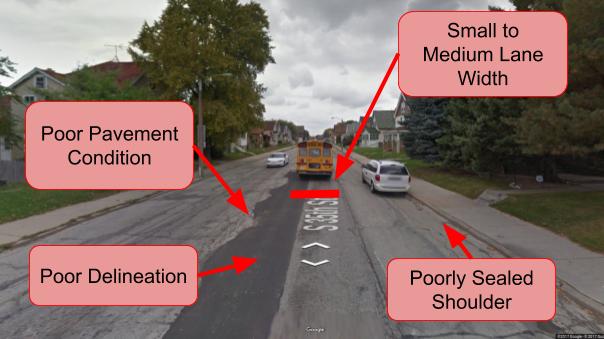}
  \end{subfigure}

  \begin{subfigure}{\linewidth}
    \centering
    \includegraphics[width=\linewidth]{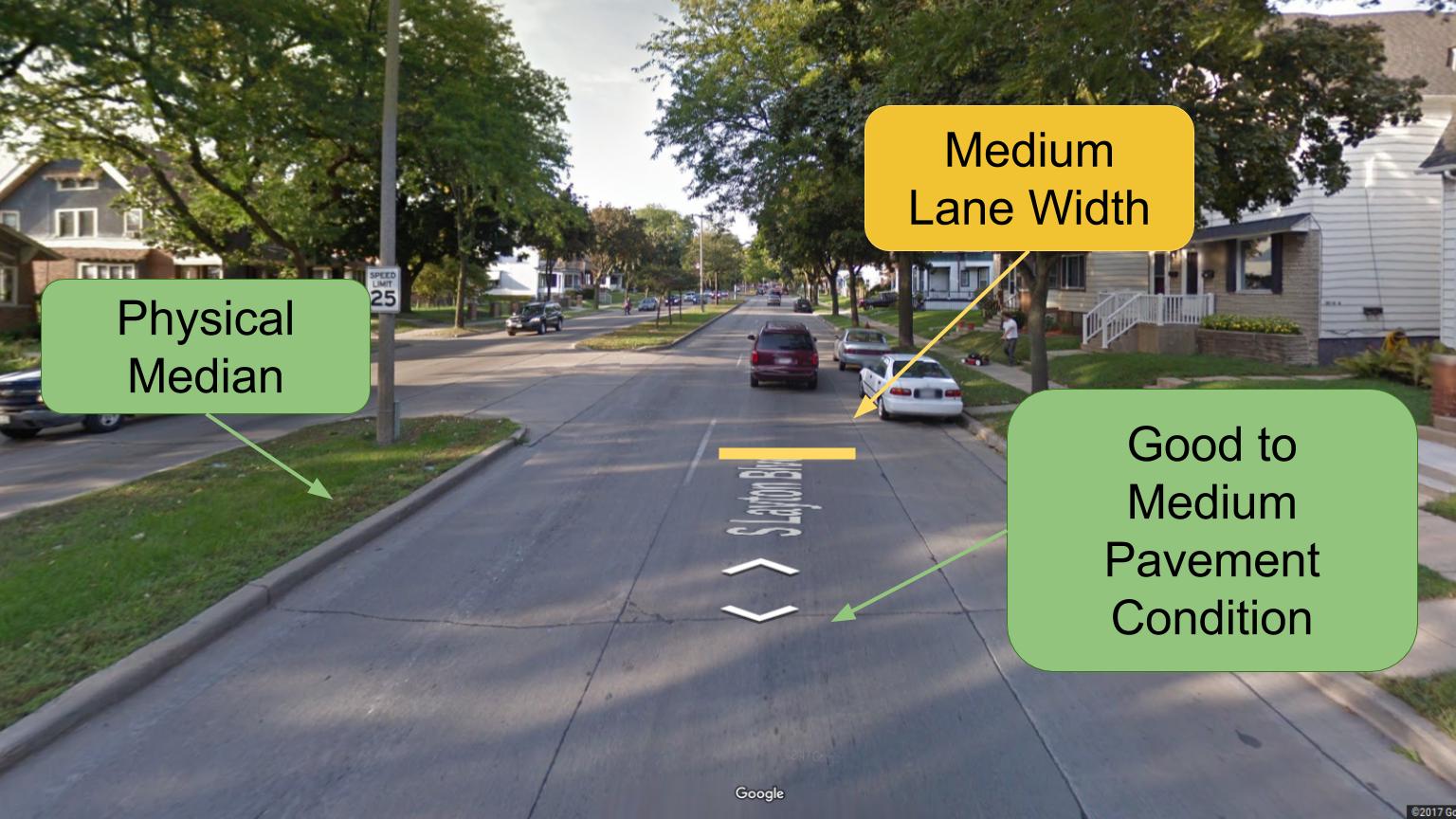}
  \end{subfigure}

  \caption{Annotation of one star (top) and four star (bottom) roads.}

  \label{annotations}

\end{figure}

The two routes depicted in Figure~\ref{routes} demonstrate the impact
of safety-aware routing in a major US urban area.
Figure~\ref{routes}-right shows a less risky, but longer, route chosen by
the enhanced routing engine, while Figure~\ref{routes}-left shows the
default route. The enhanced routing engine chooses this route to
circumvent a low (1 or 2) safety scoring road and instead travel on a
high (4 or 5) scoring road.  Figure~\ref{annotations}-top shows a
panorama from the higher risk road.  It has numerous issues, including
poor pavement condition and small lane widths.
Figure~\ref{annotations}-bottom shows a panorama from the less risky road.
This road has wider lanes, better pavement, and a physical median.
While this route may take longer, it clearly traverses a less risky path.

While this is a minor modification to an existing routing engine, we
think the ability to optimize for safety over speed could lead to
significant reduction in injuries and deaths for vehicle users. 

\section{Conclusion}

In this paper, we introduced an automated approach to estimate roadway
safety that is significantly faster than previous methods requiring
significant manual effort. We demonstrate how a combination of a
spatial attention mechanism, transfer learning, multi-task learning,
and unsupervised learning results in the best performance on a
large dataset of real-world images. This approach has the potential to
dramatically affect the deployment of limited transportation network
improvement funds toward locations that will have maximum impact on
safety. The outlined approach addresses the main concern of many
agencies in deploying systemic analysis (such as usRAP)---cost to
collect and process data.  As the availability of street-level
panoramas is growing rapidly, employing automation techniques could
allow many more agencies to take advantage of systemic road safety
techniques.  For many agencies, there are few options available to
prioritize roadway safety investments. Even when and where sufficient
data are available, smaller agencies typically lack the expertise to
conduct robust safety analysis as described in the HSM
\cite{national2010highway} for high crash location assessment or in usRAP for
systemic study. Our approach could reduce the cost of systemic analysis to
these agencies, or help larger agencies assess more of their roads, more
frequently. For future contributions, we plan to explore at least three
offshoots of this work: 1) apply the proposed method to additional highway
safety tasks, 2) integrate overhead imagery, and 3) make assessments using
multiple panoramas.

\section*{Acknowledgments}

We gratefully acknowledge the financial support of NSF CAREER grant
IIS-1553116 and computing resources provided by the
University of Kentucky Center for Computational Sciences, including a
hardware donation from IBM.

{\small
\bibliographystyle{ieee}
\bibliography{main}

\begin{thebibliography}{10}\itemsep=-1pt

\bibitem{usrap}
{United States Road Assessment Program}.
\newblock \url{http://www.usrap.org}.
\newblock Accessed: 2018-1-13.

\bibitem{arietta2014city}
S.~M. Arietta, A.~A. Efros, R.~Ramamoorthi, and M.~Agrawala.
\newblock City forensics: Using visual elements to predict non-visual city
  attributes.
\newblock {\em IEEE Transactions on Visualization and Computer Graphics},
  20(12):2624--2633, 2014.

\bibitem{cordts2016cityscapes}
M.~Cordts, M.~Omran, S.~Ramos, T.~Rehfeld, M.~Enzweiler, R.~Benenson,
  U.~Franke, S.~Roth, and B.~Schiele.
\newblock The cityscapes dataset for semantic urban scene understanding.
\newblock In {\em IEEE Conference on Computer Vision and Pattern Recognition},
  2016.

\bibitem{dubey2016deep}
A.~Dubey, N.~Naik, D.~Parikh, R.~Raskar, and C.~A. Hidalgo.
\newblock Deep learning the city: Quantifying urban perception at a global
  scale.
\newblock In {\em European Conference on Computer Vision}, 2016.

\bibitem{geiger2012we}
A.~Geiger, P.~Lenz, and R.~Urtasun.
\newblock Are we ready for autonomous driving? the kitti vision benchmark
  suite.
\newblock In {\em IEEE Conference on Computer Vision and Pattern Recognition},
  2012.

\bibitem{gibson2017analysis}
B.~Gibson.
\newblock Analysis of autonomous vehicle policies.
\newblock Technical report, Kentucky Transportation Center, 2017.

\bibitem{harwood2010validation}
D.~Harwood, K.~Bauer, D.~Gilmore, R.~Souleyrette, and Z.~Hans.
\newblock Validation of us road assessment program star rating protocol:
  Application to safety management of us roads.
\newblock {\em Transportation Research Record: Journal of the Transportation
  Research Board}, 2147:33--41, 2010.

\bibitem{he2015delving}
K.~He, X.~Zhang, S.~Ren, and J.~Sun.
\newblock Delving deep into rectifiers: Surpassing human-level performance on
  imagenet classification.
\newblock In {\em IEEE International Conference on Computer Vision}, 2015.

\bibitem{He2014IntelligentCR}
W.~He, K.~Hwang, and D.~Li.
\newblock Intelligent carpool routing for urban ridesharing by mining gps
  trajectories.
\newblock {\em IEEE Transactions on Intelligent Transportation Systems},
  15:2286--2296, 2014.

\bibitem{national2010highway}
{Joint Task Force on the Highway Safety Manual}.
\newblock {\em Highway Safety Manual}, volume~1.
\newblock American Association of State Highway and Transportation Officials,
  2010.

\bibitem{kingma2014adam}
D.~Kingma and J.~Ba.
\newblock Adam: A method for stochastic optimization.
\newblock In {\em International Conference on Learning Representations}, 2014.

\bibitem{krizhevsky2012imagenet}
A.~Krizhevsky, I.~Sutskever, and G.~Hinton.
\newblock Imagenet classification with deep convolutional neural networks.
\newblock In {\em Advances in Neural Information Processing Systems}, 2012.

\bibitem{lee2015predicting}
S.~Lee, H.~Zhang, and D.~J. Crandall.
\newblock Predicting geo-informative attributes in large-scale image
  collections using convolutional neural networks.
\newblock In {\em IEEE Winter Conference on Applications of Computer Vision},
  2015.

\bibitem{lin2014network}
M.~Lin, Q.~Chen, and S.~Yan.
\newblock Network in network.
\newblock In {\em International Conference on Learning Representations}, 2014.

\bibitem{lin2015learning}
T.-Y. Lin, Y.~Cui, S.~Belongie, and J.~Hays.
\newblock Learning deep representations for ground-to-aerial geolocalization.
\newblock In {\em IEEE Conference on Computer Vision and Pattern Recognition},
  2015.

\bibitem{mattyus2016hd}
G.~M{\'a}ttyus, S.~Wang, S.~Fidler, and R.~Urtasun.
\newblock Hd maps: Fine-grained road segmentation by parsing ground and aerial
  images.
\newblock In {\em IEEE Conference on Computer Vision and Pattern Recognition},
  2016.

\bibitem{naik2014streetscore}
N.~Naik, J.~Philipoom, R.~Raskar, and C.~Hidalgo.
\newblock Streetscore-predicting the perceived safety of one million
  streetscapes.
\newblock In {\em IEEE Conference on Computer Vision and Pattern Recognition
  Workshops}, 2014.

\bibitem{nye2017usrap}
B.~Nye, E.~Fitzsimmons, and S.~Dissanayake.
\newblock {Demonstration of the United States Road Assessment (usRAP) as a
  Systematic Safety Tool for Two Lane Roadways and Highways in Kansas}.
\newblock {\em Journal of the Transportation Research Forum}, 56(1), 2017.

\bibitem{ordonez2014learning}
V.~Ordonez and T.~L. Berg.
\newblock Learning high-level judgments of urban perception.
\newblock In {\em European Conference on Computer Vision}, 2014.

\bibitem{DBLP:journals/corr/QuerciaSA14}
D.~Quercia, R.~Schifanella, and L.~M. Aiello.
\newblock The shortest path to happiness: Recommending beautiful, quiet, and
  happy routes in the city.
\newblock {\em CoRR}, abs/1407.1031, 2014.

\bibitem{russakovsky2015imagenet}
O.~Russakovsky, J.~Deng, H.~Su, J.~Krause, S.~Satheesh, S.~Ma, Z.~Huang,
  A.~Karpathy, A.~Khosla, M.~Bernstein, et~al.
\newblock Imagenet large scale visual recognition challenge.
\newblock {\em International Journal of Computer Vision}, 115(3):211--252,
  2015.

\bibitem{salesses2013collaborative}
P.~Salesses, K.~Schechtner, and C.~A. Hidalgo.
\newblock The collaborative image of the city: mapping the inequality of urban
  perception.
\newblock {\em PloS one}, 8(7):e68400, 2013.

\bibitem{Sha:2013:SVN:2444776.2444798}
W.~Sha, D.~Kwak, B.~Nath, and L.~Iftode.
\newblock Social vehicle navigation: Integrating shared driving experience into
  vehicle navigation.
\newblock In {\em Proceedings of the 14th Workshop on Mobile Computing Systems
  and Applications}, 2013.

\bibitem{simonyan2014vgg}
K.~Simonyan and A.~Zisserman.
\newblock Very deep convolutional networks for large-scale image recognition.
\newblock In {\em International Conference on Learning Representations}, 2015.

\bibitem{wang2017torontocity}
S.~Wang, M.~Bai, G.~Mattyus, H.~Chu, W.~Luo, B.~Yang, J.~Liang, J.~Cheverie,
  S.~Fidler, and R.~Urtasun.
\newblock Torontocity: Seeing the world with a million eyes.
\newblock In {\em IEEE International Conference on Computer Vision}, 2017.

\bibitem{workman2015deepfocal}
S.~Workman, C.~Greenwell, M.~Zhai, R.~Baltenberger, and N.~Jacobs.
\newblock {DeepFocal: A Method for Direct Focal Length Estimation}.
\newblock In {\em International Conference on Image Processing}, 2015.

\bibitem{workman2017natural}
S.~Workman, R.~Souvenir, and N.~Jacobs.
\newblock {Understanding and Mapping Natural Beauty}.
\newblock In {\em IEEE International Conference on Computer Vision}, 2017.

\bibitem{workman2017unified}
S.~Workman, M.~Zhai, D.~J. Crandall, and N.~Jacobs.
\newblock A unified model for near and remote sensing.
\newblock In {\em IEEE International Conference on Computer Vision}, 2017.

\bibitem{wu2012evaluation}
K.-F. Wu, S.~Himes, and M.~Pietrucha.
\newblock Evaluation of effectiveness of the federal highway safety improvement
  program.
\newblock {\em Transportation Research Record: Journal of the Transportation
  Research Board}, (2318):23--34, 2012.

\bibitem{xu2016deaths}
J.~Xu, K.~D. Kochanek, S.~L. Murphy, and B.~Tejada-Vera.
\newblock Deaths: final data for 2014.
\newblock {\em National Vital Statistics Reports}, 65(5):1--122, 2016.

\bibitem{zhai2016context}
M.~Zhai, S.~Workman, and N.~Jacobs.
\newblock {Detecting Vanishing Points using Global Image Context in a
  Non-Manhattan World}.
\newblock In {\em IEEE Conference on Computer Vision and Pattern Recognition},
  2016.

\bibitem{zhou2017places}
B.~Zhou, A.~Lapedriza, A.~Khosla, A.~Oliva, and A.~Torralba.
\newblock Places: A 10 million image database for scene recognition.
\newblock {\em IEEE Transactions on Pattern Analysis and Machine Intelligence},
  PP(99), 2017.

\bibitem{zhou2014learning}
B.~Zhou, A.~Lapedriza, J.~Xiao, A.~Torralba, and A.~Oliva.
\newblock Learning deep features for scene recognition using places database.
\newblock In {\em Advances in Neural Information Processing Systems}, 2014.

\end{thebibliography}
}

\end{document}